# DeepFood: Deep Learning-based Food Image Recognition for Computer-aided Dietary Assessment

Chang Liu[1], Yu Cao[1], Yan Luo[1], Guanling Chen[1], Vinod Vokkarane[1], Yunsheng Ma[2],

[1] The University of Massachusetts Lowell, One University Ave, Lowell, MA, 01854, USA
ycao@cs.uml.edu
[2] The University of Massachusetts Medical School, 419 Belmont Street, Worcester, MA, 01605, USA
Yunsheng.Ma@umassmed.edu

**Abstract.** Worldwide, in 2014, more than 1.9 billion adults, 18 years and older, were overweight. Of these, over 600 million were obese. Accurately documenting dietary caloric intake is crucial to manage weight loss, but also presents challenges because most of the current methods for dietary assessment must rely on memory to recall foods eaten. The ultimate goal of our research is to develop computer-aided technical solutions to enhance and improve the accuracy of current measurements of dietary intake. Our proposed system in this paper aims to improve the accuracy of dietary assessment by analyzing the food images captured by mobile devices (e.g., smartphone). The key technique innovation in this paper is the deep learning-based food image recognition algorithms. Substantial research has demonstrated that digital imaging accurately estimates dietary intake in many environments and it has many advantages over other methods. However, how to derive the food information (e.g., food type and portion size) from food image effectively and efficiently remains a challenging and open research problem. We propose a new Convolutional Neural Network (CNN)-based food image recognition algorithm to address this problem. We applied our proposed approach to two real-world food image data sets (UEC-256 and Food-101) and achieved impressive results. To the best of our knowledge, these results outperformed all other reported work using these two data sets. Our experiments have demonstrated that the proposed approach is a promising solution for addressing the food image recognition problem. Our future work includes further improving the performance of the algorithms and integrating our system into a real-world mobile and cloud computing-based system to enhance the accuracy of current measurements of dietary intake.

## 1 Introduction

Accurate estimation of dietary caloric intake is important for assessing the effectiveness of weight loss interventions. Current methods for dietary assessment rely on self-report and manually recorded instruments (e.g., 24-hour dietary recall [1] and food frequency questionnaires [2]). Though the 24-hour dietary recall is the gold standard for reporting, this method still experiences bias as the participant is required to estimate their dietary intake (short and long term). Assessment of dietary intake by the participant can result in underreporting and underestimating of food intake [3, 4]. In order to reduce participant bias and increase the accuracy of self-report, enhancements are needed to supplement the current dietary recalls. One of the potential solutions is a mobile cloud computing system, which is to employ mobile computing devices (e.g., smartphone) to capture the dietary information in natural living environments and to employ the computing capacity in the cloud to analyze the dietary information automatically for objective dietary assessment [5-15]. Among the large selection of mobile cloud computing software for health, many have proposed to improve dietary estimates [13-15]. While these apps have features to track food intake, exercise, and save data in the cloud, the user has to manually enter all their information. To overcome these barriers, some research and development efforts have been made over the last few years for visual-based dietary information analysis [5-12]. While progresses have been made, how to derive the food information (e.g., food type) from food image effectively and efficiently remains a challenging and open research problem.

In this paper, we propose new deep learning-based [16] food image recognition algorithm to address this challenge. The proposed approach is based on Convolutional Neural Network (CNN) with a few major optimizations. The experimental results of applying the proposed approach to two real-world datasets have demonstrated the effectiveness of our solution.

The rest of the paper is organized as follows. Section 2 introduces related work in computer-aided dietary assessment and visual-based food recognition. Section 3 presents the proposed deep learning-based approach for food image recognition. Section 4 describes the implementation details and the evaluation results of our proposed algorithms. We make concluding remarks in Section 5.

## 2    Related Work

The first related research area is technology solutions for enhancing the accuracy of dietary measurement. As we have introduced before, the ubiquitous nature of mobile cloud computing invites an unprecedented opportunity to discover early predictors and novel biomarkers to support and enable smart care decision making in connection with health scenarios, including that of dietary assessment. There are thousands of mobile cloud health software (e.g., mobile health Apps available for iPhone, iPad, and Android) and many mobile health hardware options (e.g., activity tracker, wireless heart rate monitors). Among this huge selection, many have proposed to improve dietary estimates [13-15]. While these Apps have features to track food intake, exercise, and save data in the cloud, the user still has to manually enter everything they ate. Several apps have an improved level of automation. For example, Meal Snap [17] estimates the calorie content by asking the user to take a picture, dial in data such as whether you are eating breakfast or lunch, and add a quick text label. However, the accuracy of calorie estimation is unstable and is heavily dependent upon the accuracy of manually entered text input from users. Another App named "Eatly" [18] simply rates the food into one of the three categories ("very healthy", "it's O.K.", and "unhealthy") using the food image taken by the user. However, the rating is actually manually performed by the app's community of users, instead of by automated computer algorithms.

The second related research area is visual-based dietary information analysis [5-12]. Yang et al. [6] proposed a method to recognize fast food using the relative spatial relationships of local features of the ingredients followed by a feature fusion method. This method only works for a small number of food categories (61 foods) and is difficult to extend to composite or homemade food. Matsuda et al. [7] proposed an approach for multiple food recognition using a manifold ranking-based approach and co-occurrence statistics between food items, which were combined to address the multiple food recognition issue. However, this type of solution is computationally intensive and may not be practical for deployment within the mobile cloud-computing platform. A sequence of papers [8-10] from Purdue University TADA project [11] covered food item identification, food volume estimation, as well as other aspects of dietary assessment, such as mobile interface design and food image database development. The majority of their techniques for food recognition are based on traditional signal processing techniques with hand-engineered features. Recently, due to the occurrence of large annotated dataset like ImageNet [19], Microsoft COCO [20], and the development of powerful machine equipped with GPU, it is plausible to train large and complex CNN models for accurate recognition, which surpassed most of the methods adopted using hand-crafted features [21]. In this paper, we employ machine-learned features with deep learning based method, rather than the hand engineered features, to achieve a much higher accuracy.

## 3 Proposed Approach

In this paper, we propose a new deep learning-based approach to address the food image recognition problem. Specifically, we propose Convolutional Neural Network (CNN)-based algorithms with a few major optimizations, such as an optimized model and an optimized convolution technique. In the following sub-sections, we will first introduce the background and motivations of the proposed approach, follows with the detailed introduction of the proposed approach.

### 3.1 Deep Learning, Convolutional Neural Network (CNN), and Their Applications to Visual-based Food Image Recognition

Deep learning [16, 22], aims to learn multiple levels of representation and abstraction that help infer knowledge from data such as images, videos, audio, and text, is making astonishing gains in computer vision, speech recognition, multimedia analysis, and drug designing [23]. Briefly speaking, there are two main classes of deep learning techniques: purely supervised learning algorithms (e.g., Deep Convolutional Network [21]), unsupervised and semi-supervised learning algorithms (e.g., Denoising Autoencoders [24], Restricted Boltzmann Machines, and Deep

Boltzmann Machines [25]). Our proposed approach belongs to the first category (supervised learning algorithms). With the help of large-scale and well-annotated dataset like ImageNet [19], it's now feasible to perform large scale supervised learning using Convolutional Neural Network(CNN). The issue of convergence has been addressed by Hinton's work in 2006 [22]. Subsequent theoretical proof and experimental results both shows that large scale pre-trained models in large domain, with specific small scale unlabeled data in another domain, will give excellent result in image recognition and object detection [26]. To address the issue of limited abilities of feature representation, many researchers have proposed more complex CNN network structure, like VGG [27], ZFNet [28], GoogLeNet [29] and so on. On the other hand, ReLU [30] is also proposed to make it converge faster and also gains a better accuracy. Most of current researchers have put efforts in making the network deeper and avoid saturation problem [21, 27, 31].

Inspired by the advances of deep learning technique, some researchers have applied deep learning for visual-based food image recognition. In a paper by Kawano et al [32], the researchers developed an Android application to collect and label food image. They also created a food image database named "UEC-256 food image data set". With this data set, they first conducted some experiments using SIFT features and SVM, and shows much better result comparing with PFID [33]. Then, they used AlexNet [21] to the same data sets and showed much better result than SIFT-SVM-based method [34].

### 3.2 Proposed CNN-based Approach for Visual-based Food Image Recognition

Our proposed approach was directly inspired by and rooted from LeNet-5 [35], AlexNet [21], and GoogLeNet [29]. The original idea of CNN was inspired by the neuroscience model of primate visual cortex [36]. The key insights from paper [36] is how to make the machine learning with multiple level neurons like human mind. In human brain, it's known that different neurons control different perception functionality, how to make the computer recognize and think in human-like way has long been a topic for many artificial intelligence experts. In [35] the article by LeCun et al., they proposed the initial structure of LeNet-5, which is considered to the first successful trial in deep learning. In their paper, a 7-layer network structure is proposed to represent human-written digital characters and used for digits recognition. The input of the network is 32x32 grey-scale image, after several layers of convolution and sub-sampling, a feature map is generated and feed into the two fully-connected layers. After the fully-connected layer's computation, a 10-class output is generated, representing the digital 0 to 9.

This network shows the basic components of convolutional neural networks (CNN). It's consisted of three convolutional layers marked as C1, C3 and C5, sub-sampling layers marked as S2, S4 and fully connected layers as F6 and output layer. For convolutional layer, a receptive field (we call it fixed-size patch or kernel) is chosen to compute convolution with the same size patch in the input. A stride is set to make sure every pixel in the original image or feature map is covered and generates the corresponding output in the output feature map. After the operation of convolution, a sub-sampling is done within the feature map to reduce the dimension and avoid repeat computation. Finally, fully connected layers are used to concatenate the multi-dimension feature maps and to map the feature into fix-size category as a classifier. All these layers have trainable parameters (weights) adjusted when it's training using character sample images. According to some of the latest research, researchers are putting more efforts in strengthening the capabilities of the representing image features using more complex model. In the article [21], a 7-layer model called AlexNet, consisting 5 convolutional layers and 2 fully-connected layers is used with large scale labeled image dataset ImageNet. Since then, more and more work is done to increase the number of layers and layer size, while using Dropout, ReLU to address the problem of overfitting and saturating. In the following years, ZFNet, VGG, GoogLeNet are developed using more complex neurons, computation units and layer structures.

Similar to GoogleNet, we employed an Inception module to increase the representation power of neural network. This work is motivated by the Network-in-Network approach proposed by Lin et al [37]. In this module, additional 1x1 convolutional layers are added to the network, increasing the depth of overall network structure. On the other hand, this additional module can reduce the dimension of feature map, thus removing the computation bottlenecks. Normally, an Inception module takes the feature map as the input, followed with several convolutional layers varying from *1x1* convolutions, to *3x3* and *5x5* convolutions, and max-pooling layers like *3x3* pooling. Each layer generates different output and then these filters are concatenated into one feature map as the output. The outputs of the Inception module are used for next layer's convolution or pooling.

Based on the aforementioned inception modules, an optimized convolution is used to conduct dimension reduction and depth increasing. The input is not fed directly into the *3x3* and *5x5* convolutional layer. Instead, an additional *1x1* convolutional layer is added to reduce the input dimension. Furthermore, after the *3x3* max-pooling layer, the output is fed into an additional *1x1* convolutional layer. This way the Inception module is adjusted with more depth and less dimension. Similar to GooglNet, experiment shows that this network enhances the capturing of more visual information under constrained computational complexity. The improved inception module is illustrated in Fig 1. The dotted rectangle shows the added *1x1* convolution layer. In this Figure, we used dotted convolutional layer called convLayer to represent the layers that we added in the Inception module. Unlike the previous network structure that doesn't contain the dotted layer, which is only three layers: previous layer, conv/max-pool layer and concateLayer, after adding these convLayers, we have four layers: previous layer, conv/max-pool layer, conv layer and concateLayer. In this way, a feature map contains much more information than before.

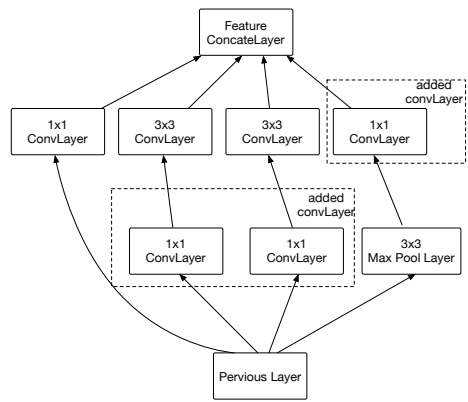

**Fig 1.** Inception Module

After the Inception module is formed, we use multiple modules to form the GoogLeNet, as shown in Fig 3, the two modules are connected via an additional max pooling layer, each module takes the input of another module, after concatenation and pooling, the output is feed into another Inception module as the input. In this way, the network forms a hierarchical level step by step. In [38], Kaiming et al. gives the general guidance for modifying models considering the influence of different depth, number of filters and filter size. In our experiment, we inherit the 22-layer network structure in GoogLeNet, run the experiment multiple times using different kernel size and stride. In our experiment, an input size of 224x224 taking RGB channels, with "*1x1*", "*3x3*" and "*5x5*" convolutions, yield the best result. Other parameters are the same as the proposed GoogLeNet,

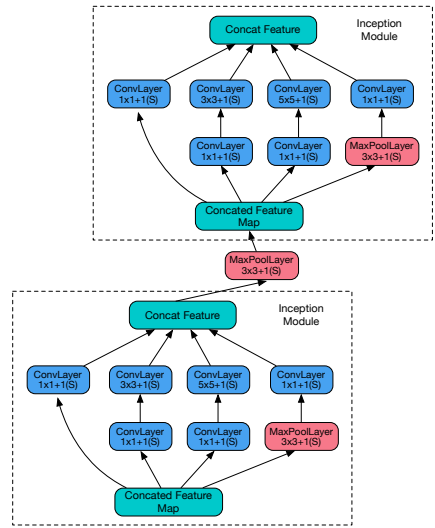

**Fig 2.** Module connection

Section 4 gives the detailed parameters when training on different data.

The network has *22* layers in depth, if only counting the layers with parameters. The average pooling layer is *5x5* filter size, and *1x1* convolutional layer is equipped with *128* filters and rectified linear activation (ReLU). In addition, the fully-connected layers are bound to *1024*-dimension for feature mapping, and it's mapped into *1000*-class output using ImageNet data set. A *70%* dropout rate is used to reduce over fitting, and the final classifier uses Softmax loss. Given different datasets, the output class number may vary according to the actual categories.

Due to the efficacy and popularity among open source community, we implement the proposed approach using Caffe [39]. In our experiment, we choose Ubuntu 14.04 as our host system. Four NVidia Tesla K40 GPUs were used to boost the training process. According to the model zoo, we used the pre-trained GoogLeNet model on ImageNet dataset with 1000 class, then fine-tuned on our own dataset UEC-256 and Food-101 with 256 classes or 101 classes. The model definition is adjusted in prototxt file in Caffe.

## 4. Experimental Results

In our experiments, we used two publicly available and challenging data sets, which were UEC-100/UEC-256 [32] and Food-101[40]. As shown in the sub-sections below, the results of our proposed approach outperformed the all the existing techniques.

### 4.1. Experimental Results on UEC-256

The first datasets we used is called UEC dataset [32], which includes two sub-data sets: UEC-100 and UEC-256 [32]. This dataset was developed by DeepFoodCam project [34]. This dataset includes a large volume of food categories with textual annotation. We note that most of the foods in the dataset are Asian foods (i.e., Japanese foods). For UEC-100, there are 100 categories with a total of 8643 images, each category has roughly 90 images. For UEC-256, there are 256 categories with a total of 28375 images; each category has roughly 110 images. All of these images are correctly label with food category and bounding-box coordinates indicating the positions of the label food partition. In our experiment, because of the requirement of large-scale training data, we chose UEC-256 as the baseline dataset. All these images are divided into 5 folds, and 3 folds were used for training while the remaining 2 folds were used for testing.

In our experiment, we first used the pre-trained model with 1000-class category from ImageNet dataset. This model is publicly available in model zoo from Caffe's community. The pre-trained model was trained using 1.2 million images for training and 100,000 images for testing. Based on the pre-trained model, we further fine-tune the model using the UEC-256 dataset whose output category number is 256. The model was fine-tuned (ft) with a base-learning rate at 0.01, a momentum of 0.9 and 100,000 iterations. The results are shown below in Table 1.

| # of Iterations | Top-1 accuracy | Top-5 accuracy |
|---|---|---|
| 4,000 | 45.0% | 76.9% |
| 16,000 | 50.4% | 78.7% |
| 32,000 | 51.2% | 79.3% |
| 48,000 | 53.1% | 80.3% |
| 64,000 | 52.5% | 80.3% |
| 72,000 | **54.7%** | **81.5%** |
| 80,000 | 53.6% | 80.1% |
| 92,000 | 54.0% | 81.0% |
| 100,000 | 53.7% | 80.7% |

**Table 2:** Comparison of accuracy on UEC-256 at different iterations using UEC-256)

We also compared our result with the original results from the DeepFoodCam papers [32, 34]. To make a fair comparison, we used the same dataset as original papers, which is UEC-100, as well as the same strategy of dividing image dataset, the result is shown in the Table 3. From this table, we can tell that our proposed method outperformed all existing methods using the same dataset:

| Method | top-1 | top-5 |
|---|---|---|
| SURF-BoF+ColorHistogram | 42.0% | 68.3% |
| HOG Patch-FV+Color Patch-FV | 49.7% | 77.6% |
| HOG Patch-FV+Color Patch-FV(flip) | 51.9% | 79.2% |
| MKL | 51.6% | 76.8% |
| Extended HOG Patch-FV+Color Patch-FV(flip) | 59.6% | 82.9% |
| DeepFoodCam(ft)[34] | 72.26% | 92.00% |
| **Proposed Approach in this Paper** | **76.3%** | **94.6%** |

Table 3: Comparison of accuracy between our proposed approach and existing approaches using the same data set (UEC-100)

### 4.2. Experimental Results on Food-101

The second data set we used is Food-101 by Lukas et al. [40]. This dataset consists of 101 categories and each category has around 1000 images. Among the 1,000 images, around 75% of them were used for training and the rest 25% were used for testing. There are 101,000 images in total in this dataset. However, since all these data were collected by food sharing websites, images do not contain any bounding box information indicating the food location. Each image only contains the label information indicating the food type. Most of the images are popular western food images.

The implementation of our algorithm for this dataset is similar to the one used in Section 4.1. We did adjust the parameters to fit for 101 food categories, and then used a base learning rate of 0.01, a momentum of 0.9 and (up to) 300,000 iterations. Based on the 1000-class pre-trained model on ImageNet dataset, we fine-tuned the model on Food-101 dataset, the accuracy is shown as the following table, and we have achieved a **77.4%** top-1 accuracy and **93.7%** top-5 accuracy.

| *# of Iterations* | *Top-1 accuracy* | *Top-5 accuracy* |
|---|---|---|
| 10,000 | 70.2% | 91.0% |
| 30,000 | 74.7% | 93.0% |
| 50,000 | 75.1% | 93.0% |
| 70,000 | 74.0% | 92.1% |
| 90,000 | 76.3% | 93.4% |
| 160,000 | 76.6% | 93.4% |
| 180,000 | 77.2% | 93.3% |
| 200,000 | 76.9% | 93.4% |
| 250,000 | **77.4%** | **93.7%** |
| 300,000 | 76.4% | 93.0% |

Table 4: Comparison of accuracy on Food-101 at different iterations

We also compared our experiment with the state of the art techniques using the Food-101 datasets for evaluation. As shown in Table 5, our proposed method is better than all existing work using the same dataset and division.

| *Method* | *top-1* | *top-5* |
|---|---|---|
| CNN-based Approach from Lukas et. al [40] | 56.40% | NA |
| RFDC-based Approach from Lukas et. al[40] | 50.76% | NA |
| **Proposed Approach in this Paper** | **77.4%** | **93.7%** |

**Table 5:** Comparison of accuracy using different method on Food-101

From the above table, we can see that pre-trained model with domain specific fine-tuning can boost the classification accuracy significantly. And fine-tuning strategy improves the accuracy comparing with non-fine-tuning method. In this table, we use "ft" to represent the method using fine-tuning, otherwise "no ft" means that we don't use fine-tuning and directly train the CNN model using designed architecture. The "NA" value in the "top-5" column means "not available", as we used the original experiment data from their paper[40], and they don't provide the top-5 result in it.

### 4.3. The Employment of Bounding Box

In the above experiment, we have shown that our proposed approach outperformed all existing approach. One further improvement is to add a pre-processing step with bounding box before fine-tuning. Specifically, based on the original image dataset, we first used the bounding box to crop the raw image. After this processing, only the food image part is remained for training and testing. We then conducted similar experiment on UEC dataset as shown Section 4.1 and 4.2.

| *Method* | *Top-1 Accuracy* | *Top-5 Accuracy* |
|---|---|---|
| Proposed Approach without Bounding Box | 54.7% | 81.5% |
| Proposed Approach with Bounding Box | **63.8%** | **87.2%** |

**Table 6:** Comparison of accuracy of proposed approach using bounding box on UEC-256

We also conducted the experiment on UEC-100, as follows:

| *Method* | *Top-1 Accuracy* | *Top-5 Accuracy* |
|---|---|---|
| Proposed Approach without Bounding Box | 57.0% | 83.4% |
| Proposed Approach with Bounding Box | **77.2%** | **94.8%** |

**Table 7:** Comparison of accuracy of proposed approach using bounding box on UEC-100

Food-101 was not used as it does not contain bounding box information and we cannot preprocess the image. Our experiment in these two subsets on UEC dataset shows that using bounding box information will significantly boost the classification accuracy. One intuitive explanation for this result is that the cropped image using bounding-box eliminates the abundant information in the raw image and forms a more accurate and clear image candidate for training, which yields a more accurate model for classification during the testing.

### 4.4. Running Time

Training a large model requires a large amount of time. On a K40 machine, it takes 2 to 3 seconds per image for forward-backward pass using GoogLeNet. Since large dataset like ImageNet, Microsoft COCO contains so many images, it's a bit waste of time to train the model from scratch. One practical strategy is to use the pre-trained model in model zoo, which is public for all researchers. In our real experiment, the training time is influenced by how powerful the machine or GPU is, how large the image candidate is, how many iterations we choose, and what value we choose for learning rate etc. According to the rough estimation, if we use the pre-trained GoogLeNet model, then fine-tune on the UEC-100, UEC-256, Food-101 dataset, it roughly takes 2 to 3 days nonstop for a server equipped with Nvidia K40 GPU to train the model. After we have trained the model, we directly apply the model for classifying the image. It takes less than 1 minute to test one image.

## 5. Conclusion

Obesity is a disorder involving excessive body fat that increases the risk of type 2 diabetes and cardiovascular diseases. In 2014, about 13% of the world's adult population (11% of men and 15% of women) were obese. Accurate estimation of dietary intake is important for assessing the effectiveness of weight loss interventions. In order to reduce bias and improve the accuracy of self-report, we proposed new algorithms to analyze the food images captured by mobile devices (e.g., smartphone). The key technique innovation in this paper is the deep learning-based food image recognition algorithms. Our proposed algorithms are based on Convolutional Neural Network (CNN). Our experimental results on two challenging data sets using our proposed approach exceed the results from all existing approaches. In the future, we plan to improve performance of the algorithms and integrate our system into a real-word mobile devices and cloud computing-based system to enhance the accuracy of current measurements of dietary caloric intake.

**Acknowledgments.** This project is supported in partial by National Science Foundation of the United States (Award No. 1547428, 1541434, 1440737, and 1229213). Points of view or opinions in this document are those of the authors and do not represent the official position or policies of the U.S. NSF.